\documentclass{article}
\usepackage{spconf,amsmath,graphicx}

\usepackage{times}  % DO NOT CHANGE THIS
\usepackage{helvet} % DO NOT CHANGE THIS
\usepackage{courier}  % DO NOT CHANGE THIS
\usepackage[hyphens]{url}  % DO NOT CHANGE THIS
\usepackage{graphicx} % DO NOT CHANGE THIS
\usepackage{graphicx}  % DO NOT CHANGE THIS

\usepackage[utf8]{inputenc} % allow utf-8 input

\usepackage{booktabs}       % professional-quality tables
\usepackage{amsfonts}       % blackboard math symbols
\usepackage{nicefrac}       % compact symbols for 1/2, etc.
\usepackage{microtype}      % microtypography

\usepackage{subfigure}
\usepackage{amsmath}
\usepackage{amsthm}

\usepackage{mathrsfs}

\usepackage{multirow}
\usepackage{paralist}

\usepackage{float}
\usepackage{booktabs}
\usepackage{multirow} 
\usepackage{amssymb}
\usepackage{graphicx}
\usepackage{subfigure} 
\usepackage{caption} 
\usepackage[linesnumbered,ruled,vlined]{algorithm2e}
\SetKwInput{KwInput}{Input}                % Set the Input
\SetKwInput{KwOutput}{Output}              % set the Output
\usepackage[colorlinks,linkcolor=red,anchorcolor=blue,citecolor=blue,pagebackref=true]{hyperref}
\title{DLDL: Dynamic Label Dictionary Learning \\via Hypergraph Regularization}
\name{Shuai Shao, Mengke Wang, Rui Xu, Yan-Jiang Wang, Bao-Di Liu
\sthanks{Corresponding Author. Thanks to the Natural Science Foundation of Shandong Province, China (Grant No. ZR2019MF073) for funding.}
}
\address{
College of Control Science and Engineering, China University of Petroleum (East China)\\
\{shuaishao, mkwang, ruixu\}@s.upc.edu.cn\\
yjwang@upc.edu.cn, thu.liubaodi@gmail.com
}

\begin{document}
%\ninept
%
\maketitle
\begin{abstract}
For classification tasks, dictionary learning based methods have attracted lots of attention in recent years.  One popular way to achieve this purpose is to introduce label information to generate a discriminative dictionary to represent samples. However, compared with traditional dictionary learning, this category of methods only achieves significant improvements in supervised learning, and has little positive influence on semi-supervised or unsupervised learning. To tackle this issue, we propose a Dynamic Label Dictionary Learning (DLDL) algorithm to generate the soft label matrix for unlabeled data. Specifically, we employ hypergraph manifold regularization to keep the relations among original data, transformed data, and soft labels consistent. We demonstrate the efficiency of the proposed DLDL approach on two remote sensing datasets.
\end{abstract}
\section{Introduction}
\label{Section: Introduction}
In recent years, dictionary learning based visual classification tasks have reached or even surpassed human beings' level. The ultimate goal of dictionary learning is to obtain an overcomplete dictionary to represent samples. Early dictionary learning based methods usually ignore the discriminative information, which is not conducive to represent the connections among different categories. Following, the label information is introduced to solve this problem. Many classical dictionary learning methods, such as LC-KSVD~\cite{jiang2013label}, FDDL~\cite{yang2011fisher}, LEDL~\cite{shao2020label}, incorporated the one-hot label matrix as the constraint term to the objective function. However, these methods only achieve significant improvements in supervised learning (all the training data has labels) tasks. For semi-supervised learning (part of training data has labels) and unsupervised learning (all the training data has no label), the influence of the introduced label information on the dictionary learning framework will be greatly reduced.
\begin{figure*}
	\begin{center}
		\includegraphics[width=1.0\linewidth]{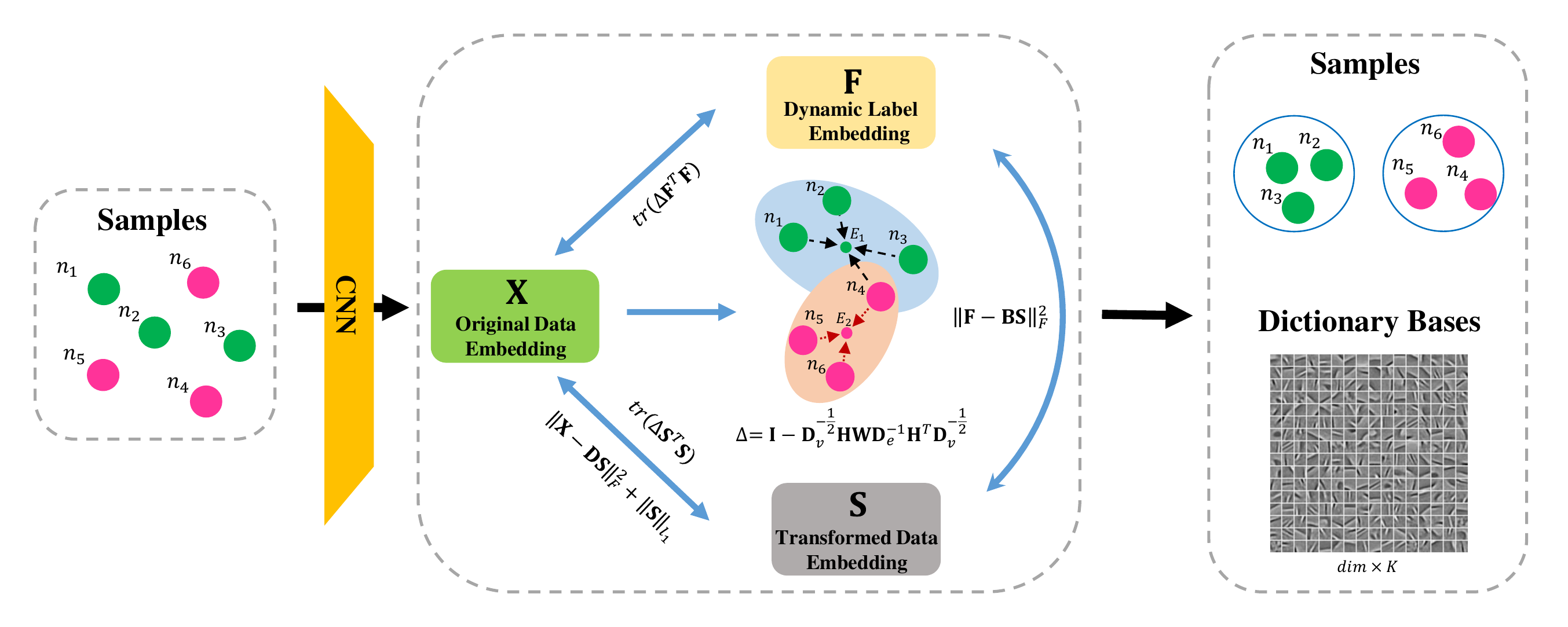}
	\end{center}
	\caption{The DLDL framework. The bule and pink backgrounds denote the hyperedge, we introduce hypergraph Laplacian operator to aggregate vertices. $E_1$ and $E_2$ represent the aggregated vertex. We construct the connections among \begin{scriptsize}$\mathbf{X}$\end{scriptsize}, \begin{scriptsize}$\mathbf{S}$\end{scriptsize} and \begin{scriptsize}$\mathbf{F}$\end{scriptsize} through Equation~\ref{equation: Objective_function_inductive}.  After that, we obtain dictionary bases to represent samples, which is helpful for classification.}
	\label{figure: DLDL}
\end{figure*}

To tackle this issue, we propose the Dynamic Label Dictionary Learning (DLDL) algorithm to dynamically produce soft labels for unlabeled training data, the soft label update with the dictionary learning.
Specifically, we introduce hypergraph manifold regularization to construct the connections among the original data, transformed data (after dictionary learning), and soft labels.
Graph/hypergraph based manifold structures have been widely applied in different fields, while in our views, most of the works can be split into two categories: $i)$ One is to build the relationship between original data and transformed data, such as HLSC~\cite{gao2012laplacian}, mHDSC~\cite{liu2014multiview}. $ii)$ Another one is to construct the connections between original data and predicted labels, including HLPN~\cite{zhang2020hypergraph}, DHSL~\cite{zhang2018dynamic} \emph{et al.}
Inspired by the two ideas,  we try to employ hypergraph manifold regularization to keep the relations among the three ones consistent, which positively influences the classification performance.

In addition, this paper purpose of classifying the remote sensing datasets. Generally, these kinds of datasets have a significant difference among different images. That is to say, there exists a more complicated relationship. However, graph-based methods are powerful ways to represent pair-wise relations for samples, but not suitable in this case. To address this problem, we introduce a hypergraph manifold structure to finish this job. Compared with the graph, hypergraph consists of vertex set and hyperedge set. Each hyperedge includes a flexible number of verices. The structure is capable of modeling the high-order relationship mentioned above. Notably, a hypergraph is the same as a simple graph when the degree of each hyperedge is restricted to $2$. Our proposed DLDL performs well in remote sensing datasets, but it is also applicable to regular datasets. We show the framework of DLDL in Figure~\ref{figure: DLDL}. 

In summary, the main contributions focus on: 
\begin{itemize}
\item 
We propose Dynamic Label Dictionary Learning to construct connections among labels, transformed data, and original data by incorporating hypergraph manifold to dictionary learning structure. We make it possible to let the label play an equally important role in supervised, semi-supervised, and unsupervised learning tasks.
\item
The proposed Dynamic Label block is a model-agnostic method, which is suitable for all subspace learning tasks.
\item
Experimental results demonstrate that the proposed DLDL significantly improves the classification performance compared with other state-of-the-art dictionary learning methods.
\end{itemize}

\section{Methodology}
\label{section: Methodology}
In this section, we introduce the details of dynamic dictionary learning method, and show the framework in Figure~\ref{figure: DLDL}. 

\subsection{Review of Dictionary Learning}
\label{section: dictionary_learning}
Our utilised datasets include training data \begin{scriptsize}$\mathbf{X} \in \mathbb{R}^{dim \times N}$\end{scriptsize} and testing data \begin{scriptsize}$\mathbf{Y}\in \mathbb{R}^{dim \times M}$\end{scriptsize},  where \begin{scriptsize}${\mathbf{x}}_i$ ($i = 1, 2, \dots$)\end{scriptsize} is the feature embedding of $i$-th sample and \begin{scriptsize}$dim$\end{scriptsize} denotes the dimension. In dictionary learning, a sparse representation \begin{scriptsize}$\mathbf{S}=[{\mathbf{s}}_1,{\mathbf{s}}_2,\dots,{\mathbf{s}}_{N}] \in{\mathbb{R}}^{K\times {N}}$\end{scriptsize} is computed over a dictionary \begin{scriptsize}$\mathbf{D}=[{\mathbf{d}}_1,{\mathbf{d}}_2,\dots,{\mathbf{d}}_K] \in{\mathbb{R}}^{dim\times K}$\end{scriptsize} by minimizing the reconstruction error, where $K$ is the number of atoms in dictionary. A general dictionary learning algorithm can be formulated as follows:
\begin{equation}
\scriptsize
\begin{split}
         &\mathop {\arg \min}\limits_{{\mathbf{D}},{\mathbf{S}}} f_1\left({\mathbf{D}},{\mathbf{S}}\right) 
         =\left\| \mathbf{X} - \mathbf{D}\mathbf{S} \right\|_F^2 + 2\alpha \left\| \mathbf{S} \right\|_{\ell_1}\\
		&{\kern 10pt}{\rm{s}}.t.{\kern 4pt}\left\| {\mathbf{d}}_{ \bullet k} \right\|_2^2 \le 1, \left( {k = 1,2, \cdots K} \right)
\end{split}
\label{equation: DL_traditional}
\end{equation}
where \begin{scriptsize}$\left\| \mathbf{X} - \mathbf{D}\mathbf{S} \right\|_F^2$\end{scriptsize} is the reconstruction error, \begin{scriptsize}${\left(  \mathbf{d}  \right)_{ \bullet k}}$\end{scriptsize} denotes the $k$-th column vector of matrix \begin{scriptsize}$\mathbf{D}$. $\left\| \mathbf{S} \right\|_{\ell_1}$\end{scriptsize} represents the sparse constraint for \begin{scriptsize}$\mathbf{S}$\end{scriptsize} (e.g. $\ell_1$-norm regularization), and $\alpha$ is a positive scalar constant. 

\subsection{Dynamic Label Generation via Hypergraph}
\label{section: Dynamic Label Generation via Hypergraph}
In this subsection, we first construct the hypergraph, then introduce Laplacian operator to generate dynamic labels.\\
\textbf{Hypergraph Construction} 
For any hypergraph based applications, a suitable hypergraph structure is necessary. Different from graph structure, hypergraph can capture high-order relations among samples. We define hypergraph as \begin{scriptsize}$\mathcal{G}=(\mathcal{V},\mathcal{E},\mathbf{W})$\end{scriptsize}, where \begin{scriptsize}$\mathcal{V}$\end{scriptsize} denotes the vertex set, each vertex denotes a sample, \begin{scriptsize}$\mathcal{E}$\end{scriptsize} is the hyperedge set, and \begin{scriptsize}$\mathbf{W}$\end{scriptsize} denotes a weight matrix of hyperedge, which is composed of diagonal elements, each element denotes the weight of the corresponding hyperedge. The connection of hyperedges and vertices can be represented by the incidence matrix \begin{scriptsize}$\mathbf{H}\in{\mathbb{R}}^{|\mathcal{V}|\times{|\mathcal{E}|}}$\end{scriptsize}. The elements in the incidence matrix are defined as follows:
\begin{equation}
\scriptsize
\begin{split}
        \mathbf{H}=
        \left\{\begin{array}{cc}
            {\exp \left(-dis\left(v, v_{c}\right)^{2}\right)} & {\text { if } v \in e} \\
            {0} & {\text{ o.w. }}
        \end{array}\right.
\end{split}
\label{equation: elements_in_H}
\end{equation}
where $e$ is one hyperedge among \begin{scriptsize}$\mathcal{E}$\end{scriptsize}, $v$ denotes a vertex in \begin{scriptsize}$\mathcal{V}$\end{scriptsize} and $v_c$ is the centroid vertex in $e$. $dis$ denotes the operator to compute the distance with knn. Besides, we define two diagonal matrices as \begin{scriptsize}$\mathbf{D}_v$\end{scriptsize} (vertex degree matrix) and \begin{scriptsize}$\mathbf{D}_e$\end{scriptsize} (hyperedge degree matrix), which are formulated as follows:
\begin{equation}
\scriptsize
\begin{split}
        \delta({e})= \sum_{{v} \in \mathcal{V}} \mathbf{H}({v},{e})
\end{split}
\label{equation: vertex_degree}
\end{equation}
\begin{equation}
\scriptsize
\begin{split}
        d({v})= \sum_{{e} \in \mathcal{E}} \mathbf{W}({e} ) \mathbf{H}({v},{e})
\end{split}
\label{equation: hyperedge_degree}
\end{equation}\\
% \begin{equation}
% \scriptsize
% \begin{split}
%         \left\{\begin{array}{ll}
%             \delta({e})= \sum_{{v} \in \mathcal{V}} \mathbf{H}({v},{e}) \\
%             d({v})= \sum_{{e} \in \mathcal{E}} \mathbf{W}({e} ) \mathbf{H}({v},{e})
%         \end{array}\right.
% \end{split}
% \label{equation: vertex_hyperedge_degree}
% \end{equation}\\
\textbf{Dynamic Label Generation}
Assume parts of training data have labels, define initial label embedding matrix as \begin{scriptsize}$\mathbf{O} \in \mathbb{R}^{C \times N}$\end{scriptsize}, where \begin{scriptsize}$C$\end{scriptsize} denotes the total number of classes. For labeled samples, \begin{scriptsize}$\mathbf{O}_{ij}$\end{scriptsize} is $1$ if the $j$-th sample belongs to the $i$-th class, and it is $0$ otherwise. For unlabeled samples, we set all elements to $0.5$. A suitable label matrix is a good guidance to learn dictionary bases. However, in \begin{scriptsize}$\mathbf{O}$\end{scriptsize}, only labeled samples own the correct label embedding, it must interfere with the updating of \begin{scriptsize}$\mathbf{D}$\end{scriptsize} and \begin{scriptsize}$\mathbf{S}$\end{scriptsize}. To tackle this problem, we propose a method to generate the dynamic label projection matrix \begin{scriptsize}$\mathbf{F} \in \mathbb{R}^{C \times N}$\end{scriptsize}, and embed it into dictionary learning. 

It propagates dynamic label information with the joint learning of the incidence matrix \begin{scriptsize}$\mathbf{H}$\end{scriptsize}, label projection matrix \begin{scriptsize}$\mathbf{F}$\end{scriptsize}, and transformed feature embedding \begin{scriptsize}$\mathbf{S}$\end{scriptsize}.
We formulate the relationship between \begin{scriptsize}$\mathbf{X}$\end{scriptsize} and \begin{scriptsize}$\mathbf{F}$\end{scriptsize} as follows:
\begin{equation}
    \scriptsize
    \begin{split}
    &f_2\left(\mathbf{F}\right)\\
    &=\frac{1}{2}\sum_{c=1}^{C} \sum_{e \in \mathcal{E}} \sum_{u,v \in \mathcal{V}} 
    \frac{\mathbf{W}(e)\mathbf{H}(u,e)\mathbf{H}(v,e)}{\delta(e)}
    \left( \frac{\mathbf{F}\left(u,c\right)}{\sqrt{d\left( u \right)}}
    - \frac{\mathbf{F}\left(v,c\right)}{\sqrt{d\left( v \right)}} \right)^2\\
    % &=\sum_{c=1}^{C} \sum_{e \in \mathcal{E}} \sum_{u,v \in \mathcal{V}}
    % \frac{\mathbf{W}(e)\mathbf{H}(u,e)\mathbf{H}(v,e)}{\delta(e)}
    % \left( \frac{\mathbf{F}\left(u,c\right)^2}{d\left( u \right)}
    % - \frac{\mathbf{F}\left(u,c\right)\mathbf{F}\left(v,c\right)}{\sqrt{d\left( u \right)d\left( v \right)}} \right)\\
    % &=\sum_{c=1}^{C} \sum_{u \in \mathcal{V}} \mathbf{F}\left(u,c\right)^2 
    % \sum_{e \in \mathcal{E}}
    % \frac{\mathbf{W}(e)\mathbf{H}(u,e)}{d(u)}
    % \sum_{v \in \mathcal{V}} \frac{\mathbf{H}(v,e)}{\delta(e)}\\
    % &{\kern 10pt}-\sum_{e \in \mathcal{E}}\sum_{u,v \in \mathcal{V}}
    % \left( 
    % \frac{\mathbf{F}\left(u,c\right)
    % \mathbf{H}(u,e)
    % \mathbf{W}(e)
    % \mathbf{H}(v,e)
    % \mathbf{F}\left(v,c\right)}
    % {\sqrt{d\left( u \right)d\left( v \right)} \delta(e)
    % } \right )\\
    &=\text{tr}\left(
    \left(
    \mathbf{I} - \mathbf{D}_v^{-\frac{1}{2}} \mathbf{H} \mathbf{W} \mathbf{D}_e^{-1} \mathbf{H}^T \mathbf{D}_v^{-\frac{1}{2}}
    \right)
    \mathbf{F}^T \mathbf{F}
    \right)\\
    &=\text{tr}\left( \mathbf{\Delta} \mathbf{F}^T \mathbf{F}
    \right)
    \end{split}
\label{equation: X_F_connect}
\end{equation}
where \begin{scriptsize}$\mathbf{\Delta}=\mathbf{I} - \mathbf{D}_v^{-\frac{1}{2}} \mathbf{H} \mathbf{W} \mathbf{D}_e^{-1} \mathbf{H}^T \mathbf{D}_v^{-\frac{1}{2}}$\end{scriptsize} denotes the normalized hypergraph Laplacian operator. By this way, we can obtain the smooth \begin{scriptsize}$\mathbf{F}$\end{scriptsize} in label space.
Following, we construct connections between \begin{scriptsize}$\mathbf{X}$\end{scriptsize} and \begin{scriptsize}$\mathbf{S}$\end{scriptsize} as:
% Moreover, we should also preserve the hypergraph-based local structure in feature space, which is helpful to construct the overcomplete basis for dictionary learning. Similar to Equation~\ref{equation: dynamic_label_regularization}, we corporate the hypergraph manifold regularization for dictionary learning as follows:
\begin{equation}
\scriptsize
    \begin{split}
        f_3\left(\mathbf{S} \right)
        =\text{tr}\left(\mathbf{\Delta} \mathbf{S}^T \mathbf{S}\right)
    \end{split}
\label{equation: X_S_connect}
\end{equation}
In the end, we introduce the general label constraint term to construct the connections for \begin{scriptsize}$\mathbf{F}$\end{scriptsize} and \begin{scriptsize}$\mathbf{S}$\end{scriptsize}, and introduce an empirical loss for \begin{scriptsize}$\mathbf{F}$\end{scriptsize} as follows:
\begin{equation}
\scriptsize
\begin{split}
        f_4\left({\mathbf{F},\mathbf{B}},{\mathbf{S}}\right) = \left\| \mathbf{F} - \mathbf{B}\mathbf{S} \right\|_F^2 + \left\| \mathbf{F} - \mathbf{O} \right\|_F^2\\
\end{split}
\label{equation: F_S_connect}
\end{equation}
where \begin{scriptsize}$\mathbf{B} \in \mathbb{R}^{C \times K}$\end{scriptsize} is the classifier.

% At the end, we formulate the dynamic label constraint term as:
% \begin{equation}
% \scriptsize
% \begin{split}
%         &\Theta\left(\mathbf{S},\mathbf{B},\mathbf{F},\mathbf{H}\right)\\ 
%         &=\beta f_1(\mathbf{H},\mathbf{F}) +  \delta f_2(\mathbf{H},\mathbf{S}) + \beta f_3(\mathbf{F},\mathbf{B},\mathbf{S})\\
%         &=\beta \text{tr}\left( \mathbf{\Delta} \mathbf{F}^T \mathbf{F}\right)
% 		+ \delta \text{tr}\left( \mathbf{\Delta} \mathbf{S}^T \mathbf{S}\right)
% 		+ \beta \left\| \mathbf{F} - \mathbf{B}\mathbf{S} \right\|_F^2 
% 		+ \beta \left\| \mathbf{F} - \mathbf{O} \right\|_F^2\\
% 		&=\text{tr}\left( \mathbf{\Delta} \left(\beta \mathbf{F}^T \mathbf{F} +\delta \mathbf{S}^T \mathbf{S}\right)\right)
% 		+ \beta \left\| \mathbf{F} - \mathbf{B}\mathbf{S} \right\|_F^2 
% 		+ \beta \left\| \mathbf{F} - \mathbf{O} \right\|_F^2
% \end{split}
% \label{equation: dynamic_label_constraint}
% \end{equation}
% where $\beta$, $\delta$ are the balanced parameters for objective function.

\subsection{Dynamic Label Dictionary Learning}
% \textbf{Objective Function} 
We summarize the above requirements. The objective function for dynamic label dictionary learning can be written as follows:
\begin{equation}
\scriptsize
\begin{split}
        &\mathop {\arg \min}\limits_{\mathbf{D}, \mathbf{S},\mathbf{F},\mathbf{B}}     \mathcal{F}(\mathbf{D},\mathbf{S},\mathbf{F},\mathbf{B})\\
        &= f_1 \left({\mathbf{D}},{\mathbf{S}}\right) 
        + f_2 \left(\mathbf{F}\right)
        + f_3\left(\mathbf{S}\right)
        + f_4\left(\mathbf{F}, \mathbf{B}, \mathbf{S}\right)\\
		&=\left\| \mathbf{X} - \mathbf{D}\mathbf{S} \right\|_F^2 
		+ 2\alpha \left\| \mathbf{S} \right\|_{\ell_1} 
		+ \delta {\kern 2pt} \text{tr}\left( \mathbf{\Delta} \mathbf{S}^T \mathbf{S} \right) \\
		&{\kern 10pt}+ \beta {\kern 2pt} \left\| \mathbf{F} - \mathbf{B}\mathbf{S} \right\|_F^2 
		+ \beta {\kern 2pt} \left\| \mathbf{F} - \mathbf{O} \right\|_F^2
		+ \beta {\kern 2pt} \text{tr}\left( \mathbf{\Delta} \mathbf{F}^T \mathbf{F} \right)\\
		&{\kern 5pt}{\rm{s}}.t.\left\| {{{\bf{d}}_{ \bullet k}}} \right\|_2^2 \le 1,
		{\kern 5pt}\left\| {{{\bf{b}}_{ \bullet k}}} \right\|_2^2 \le 1 {\kern 4pt} \left( {k = 1,2, \cdots K} \right)\\
\end{split}
\label{equation: Objective_function_inductive}
\end{equation}
where $\beta$, $\delta$ are the balanced parameters for objective function. By this way, we can get the optimal \begin{scriptsize}$\mathbf{S}$\end{scriptsize}, \begin{scriptsize}$\mathbf{D}$\end{scriptsize}, \begin{scriptsize}$\mathbf{B}$\end{scriptsize} and \begin{scriptsize}$\mathbf{F}$\end{scriptsize} by alternative optimization until the loss dose not descend. Specifically, we update \begin{scriptsize}$\mathbf{S}$\end{scriptsize} with \begin{scriptsize}$\mathbf{D}$\end{scriptsize}, \begin{scriptsize}$\mathbf{B}$\end{scriptsize} and \begin{scriptsize}$\mathbf{F}$\end{scriptsize} fixed, the closed form solution of \begin{scriptsize}$\mathbf{S}$\end{scriptsize} is:
\begin{equation}
\scriptsize
\begin{split}
        \mathbf{S}_{kn}=
            {\frac{max\left(\mathcal{J},\alpha \right) + min\left(\mathcal{J},\alpha \right)}
            {\left(\mathbf{D}^T\mathbf{D}
            +\beta {\kern 2pt} \mathbf{B}^T\mathbf{B}\right)_{kk}
            +\delta\left(\mathbf{\Delta}\right)_{nn}}}
\end{split}
\label{equation: optimization_UpdateS_Skn_1}
\end{equation}
where 
\begin{equation}
\scriptsize
\begin{split}
   \mathcal{J}&=\left(\mathbf{D}^T\mathbf{X}+\beta \mathbf{B}^T \mathbf{F} \right)_{kn}
    -\delta \sum_{r=1,r\neq n}^{N} \left(\mathbf{\Delta}\right)_{nr}\mathbf{S}_{kr}\\
    &{\kern 10pt}-\sum_{l=1,l\neq k}^{K} \left(\mathbf{D}^T\mathbf{D}
    +\beta \mathbf{B}^T\mathbf{B} \right)_{kl}\mathbf{S}_{ln}
\end{split}
\label{equation: optimization_UpdateS_J}
\end{equation}
Then we introduce blockwise coordinate descent (BCD) method \cite{liu2014blockwise} to 
directly obtain \begin{scriptsize}$\mathbf{D}$\end{scriptsize} and \begin{scriptsize}$\mathbf{B}$\end{scriptsize} as:
\begin{equation}
\scriptsize
\begin{split}
            \mathbf{D}_{\bullet k} 
            = \frac{\mathbf{X}\left(\mathbf{S}_{k \bullet} \right)^T
            -\mathbf{\tilde{D}}^k \mathbf{S} \left(\mathbf{S}_{k \bullet} \right)^T}
            {\| \mathbf{X}\left(\mathbf{S}_{k \bullet} \right)^T
            -\mathbf{\tilde{D}}^k \mathbf{S} \left(\mathbf{S}_{k \bullet} \right)^T \|_2}   
\end{split}
\label{equation: optimization_UpdateD}
\end{equation}
\begin{equation}
\scriptsize
\begin{split}
            \mathbf{B}_{\bullet k} 
            = \frac{\mathbf{F}\left(\mathbf{S}_{k \bullet} \right)^T
            -\mathbf{\tilde{B}}^k \mathbf{S} \left(\mathbf{S}_{k \bullet} \right)^T}
            {\| \mathbf{F}\left(\mathbf{S}_{k \bullet} \right)^T
            -\mathbf{\tilde{B}}^k \mathbf{S} \left(\mathbf{S}_{k \bullet} \right)^T \|_2}     
\end{split}
\label{equation: optimization_UpdateB}
\end{equation}
% \begin{equation}
% \scriptsize
% \begin{split}
%         \left\{\begin{array}{cc}
%         \mathbf{B}_{\bullet k} 
%         = \frac{\mathbf{F}\left(\left(\mathbf{S}\right)_{k \bullet} \right)^T
%         -\mathbf{\tilde{B}}^k \mathbf{S} \left(\left(\mathbf{S}\right)_{k \bullet} \right)^T}
%         {\| \mathbf{F}\left(\left(\mathbf{S}\right)_{k \bullet} \right)^T
%         -\mathbf{\tilde{B}}^k \mathbf{S} \left(\left(\mathbf{S}\right)_{k \bullet} \right)^T \|_2}  
%         \\
%         \mathbf{B}_{\bullet k} 
%         = \frac{\mathbf{F}\left(\left(\mathbf{S}\right)_{k \bullet} \right)^T
%         -\mathbf{\tilde{B}}^k \mathbf{S} \left(\left(\mathbf{S}\right)_{k \bullet} \right)^T}
%         {\| \mathbf{F}\left(\left(\mathbf{S}\right)_{k \bullet} \right)^T
%         -\mathbf{\tilde{B}}^k \mathbf{S} \left(\left(\mathbf{S}\right)_{k \bullet} \right)^T \|_2}  
%         \end{array}\right.
% \end{split}
% \label{equation: elements_in_H}
% \end{equation}
where \begin{scriptsize}$ \mathbf{\tilde{D}}=
        \left\{ \begin{array}{cc}
             {\mathbf{D}_{\bullet p}}  & {p \neq{k}}  \\
             {\mathbf{0}} & {p = k}
         \end{array}\right.$\end{scriptsize}, 
         \begin{scriptsize}$ \mathbf{\tilde{B}}=
        \left\{ \begin{array}{cc}
             {\mathbf{B}_{\bullet p}}  & {p \neq{k}}  \\
             {\mathbf{0}} & {p = k}
         \end{array}\right.$\end{scriptsize}, $\mathbf{0}$ denotes zero matrix.
In the end, we update \begin{scriptsize}$\mathbf{F}$\end{scriptsize} with \begin{scriptsize}$\mathbf{S}$\end{scriptsize}, \begin{scriptsize}$\mathbf{D}$\end{scriptsize}, \begin{scriptsize}$\mathbf{B}$\end{scriptsize} fixed, the closed form solution of \begin{scriptsize}$\mathbf{F}$\end{scriptsize} is:
\begin{equation}
\scriptsize
\begin{split}
    \mathbf{F} = \beta \left( \mathbf{\Delta} + 2\beta  \mathbf{I} \right)^{-1} 
     \left( \mathbf{B}\mathbf{S} +  \mathbf{O} \right)
\end{split}
\label{equation: optimization_UpdateF}
\end{equation}

\section{Experiment}
In this section, in order to fairly evaluate the effectiveness of our DLDL, we compared it with multiple state-of-the-art dictionary learning methods on two remote sensing datasets, include UC Merced Land Use (UCM-LU) \cite{yang2010bag} and RSSCN7~\cite{zou2015deep}. 
We first introduce the experimental setup, and then report the experimental results. At last, we conduct ablation studies to analyze the DLDL method.

% \subsection{Dataset}
% The employed benchmark datasets include UC Merced Land Use \cite{yang2010bag} and RSSCN7~\cite{zou2015deep}.
% UC Merced Land Use (UCM-LU) dataset comes from the United States Geological Survey. It contains $2100$ land-use images with $21$ categories, such as forest, intersection, medium-density residential, baseball diamond, etc. For the RSSCN7 dataset, it consists of $7$ classes of images, including parking lot, forest, lake region, etc. Each category is composed of $400$ images, and all of them come from Google Earth.
% For all the datasets, we select $5$ samples from each category for training. 
% % More details of vertex and hyperedge are listed in the Table~\ref{table: dataset_split}.

\subsection{Experimental Setup}
For all the datasets, we employ standard Resnet~\cite{he2016deep} to extract feature embedding with $2,048$ dimensions. Each category has $5$ labeled samples. After that, we fix the dictionary size to $200$ and the nearest number of knn to $10$ for all datasets. The influence of dictionary size is discussed in the following section~\ref{section: ablation study}. In addition, there are three other parameters ($\alpha$, $\beta$ and $\delta$ ) need to be tuned manually. 
Here, we give our optimal setups for best performance of DLDL. Specifically, we set $\alpha=2^{-4}$, $\beta=2^{-4}$, $\delta=2^{2}$ for UCM-LU dataset, $\alpha=2^{-4}$, $\beta=2^{-10}$, $\delta=2^{6}$ for RSSCN7 dataset. For some discussions of parameters, please refer to the section~\ref{section: ablation study}. 

\subsection{Experimental Results}
\begin{table}
        \caption{Classification results}
        \label{table: classification_results}
        % \centering
        \begin{center}
            \begin{tabular}{lcc}
                \toprule
                %  \multicolumn{2}{c}{Part} \\
                % \cmidrule(r){1-2}
                % \midrule
                Methods$\backslash$Datasets                     & UCM-LU        & RSSCN7          \\
                \midrule
                SRC (TPAMI \cite{wright2009robust}, 2009)       & 80.4$\%$      & 67.1$\%$      \\
                CRC (ICCV \cite{zhang2011sparse}, 2011)         & 80.7$\%$      & 67.7$\%$      \\
                NRC (PR \cite{xu2019sparse}, 2019)              & 81.6$\%$      & 69.7$\%$      \\
                SLRC (TPAMI \cite{deng2018face}, 2018)          & 81.0$\%$      & 66.4$\%$       \\
                Euler-SRC (AAAI \cite{liu2018euler}, 2018)      & 80.9$\%$      & 69.7$\%$      \\
                LC-KSVD (TPAMI \cite{jiang2013label}, 2013)     & 79.4$\%$      & 68.0$\%$      \\
                CSDL (NC \cite{liu2016face}, 2016)              & 80.5$\%$      & 66.7$\%$      \\
                LC-PDL (IJCAI \cite{zhang2019scalable}, 2019)   & 81.2$\%$      & 69.7$\%$      \\
                FDDL (ICCV \cite{yang2011fisher}, 2011)         & 81.0$\%$      & 64.0$\%$     \\
                LEDL (NC \cite{shao2020label}, 2020)            & 80.7$\%$      & 67.9$\%$      \\
                ADDL (TNNLS \cite{zhang2018jointly}, 2018)      & 83.2$\%$      & 72.3$\%$       \\
                CDLF (SP \cite{wang2020class}, 2020)            & 81.0$\%$      & 69.6$\%$       \\
                HLSC (TPAMI \cite{gao2012laplacian}, 2012)      & 81.4$\%$      & 71.1$\%$       \\
                \midrule
                \textbf{DLDL}                                         & \bf{85.2$\%$} & \bf{72.9$\%$}  \\
                \bottomrule
            \end{tabular}        
        \end{center}
\end{table}
We compare our DLDL with several classical classification methods, include SRC \cite{wright2009robust}, CRC \cite{zhang2011sparse}, NRC \cite{xu2019sparse}, SLRC \cite{deng2018face}, Euler-SRC \cite{liu2018euler}, LC-KSVD \cite{jiang2013label}, CSDL \cite{liu2016face}, LC-PDL \cite{zhang2019scalable}, FDDL \cite{yang2011fisher}, LEDL \cite{shao2020label}, ADDL \cite{zhang2018jointly}, CDLF \cite{wang2020class}, HLSC \cite{gao2012laplacian}. We show the experimental results in Table~\ref{table: classification_results}, and have the following observations.

\textbf{$i)$} 
Obviously see that our DLDL outperforms all the other state-of-the-art methods. On the UCM-LU dataset, DLDL achieves the best performance by an improvement of at least $2\%$, and on the RSSCN7 dataset, DLDL is able to exceed other methods at least $0.6\%$.

\textbf{$ii)$} 
Compared with traditional label embedded dictionary learning methods, including LC-KSVD, LEDL, CDLF, our proposed dynamic label helps outperform them at least $4.2\%$ on the UCM-LU dataset and $3.3\%$ on the RSSCN7 dataset. For other classical dictionary learning methods, such as CSDL, LC-PDL, FDDL, ADDL, HLSC, our method do not improve much, especially on the RSSCN7 dataset, DLDL only achieves an improvement of $0.6\%$ compared with ADDL. The reason is that all the dictionary learning based methods have their highlights, while our DLDL only introduces the dynamic label to a basic dictionary learning model. In other words, our proposed method is a model-agnostic module, which can be embedded in any dictionary learning based works to promote performance.

\subsection{Ablation Study}
\label{section: ablation study}
Our DLDL has achieved outstanding performance. It is necessary to know what the factors affecting the experimental results are. For this purpose, we design two ablation studies to discuss our proposed method further.
\begin{figure}
	\begin{center}
		\includegraphics[width=1.0\linewidth]{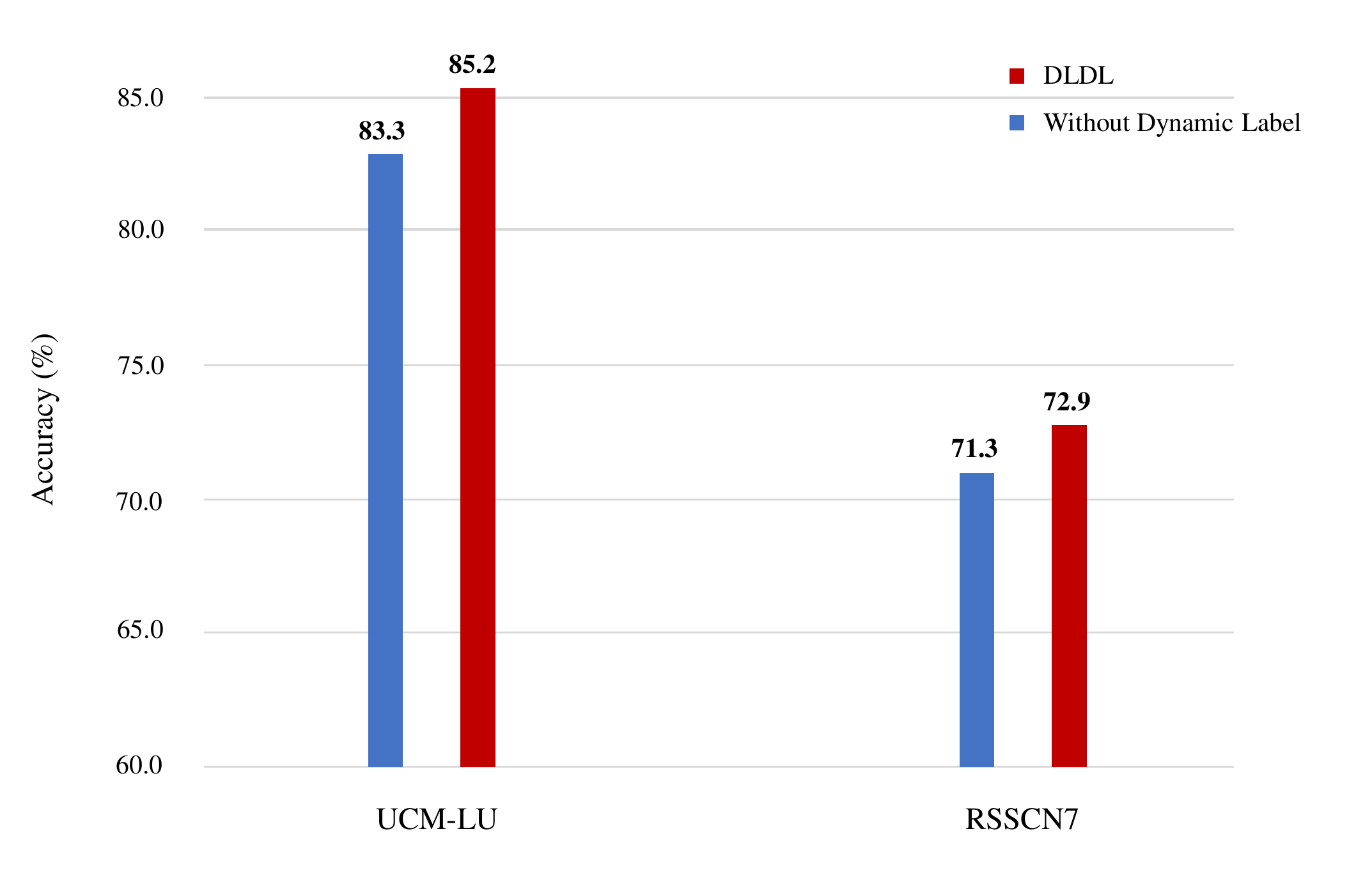}
	\end{center}
	\caption{The influence of dynamic labels.}
	\label{figure: dy-lb}
\end{figure}
\begin{figure}
    \subfigure[]{
        \begin{minipage}[t]{0.33\linewidth}
        	\begin{center}
        		\includegraphics[width=0.9\linewidth]{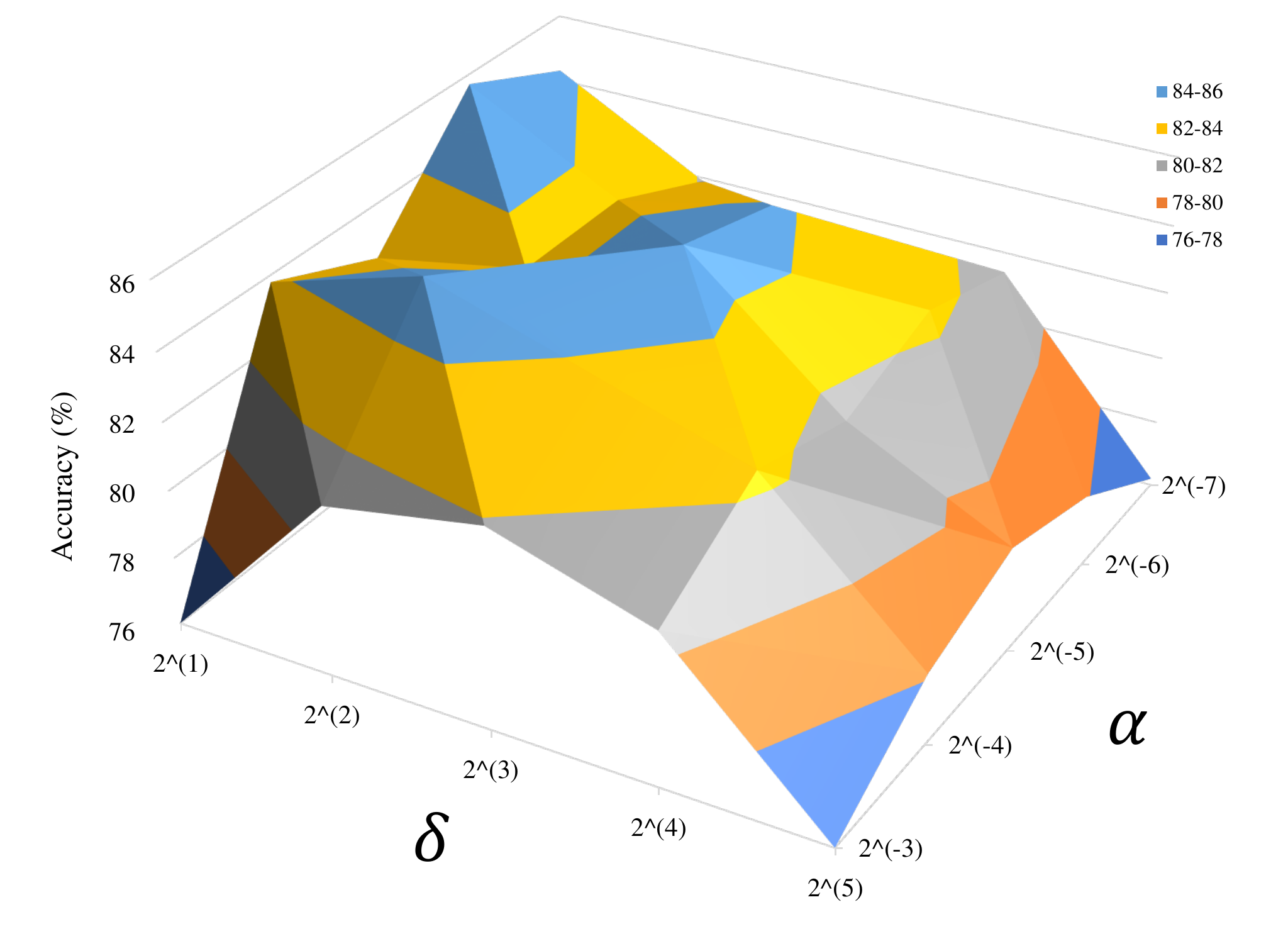}
        	\end{center}
        % 	\caption{}
        	\label{figure: alpha_beta}
        \end{minipage}%
    }%
    \subfigure[]{
        \begin{minipage}[t]{0.33\linewidth}
        	\begin{center}
        		\includegraphics[width=1\linewidth]{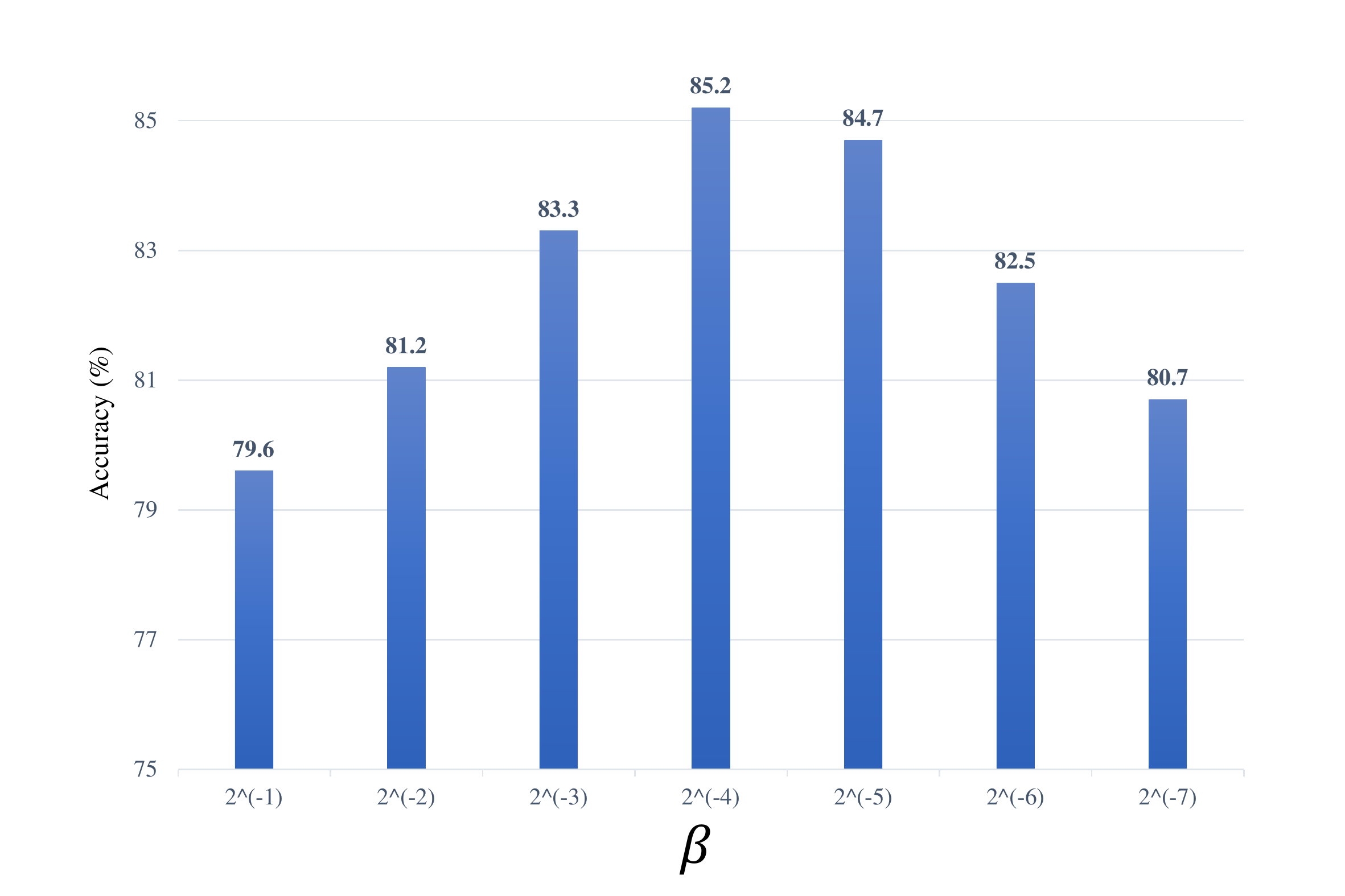}
        	\end{center}
        % 	\caption{}
        	\label{figure: epsilon}
        \end{minipage}%
    }%
    \subfigure[]{
    \begin{minipage}[t]{0.33\linewidth}
    	\begin{center}
    		\includegraphics[width=1\linewidth]{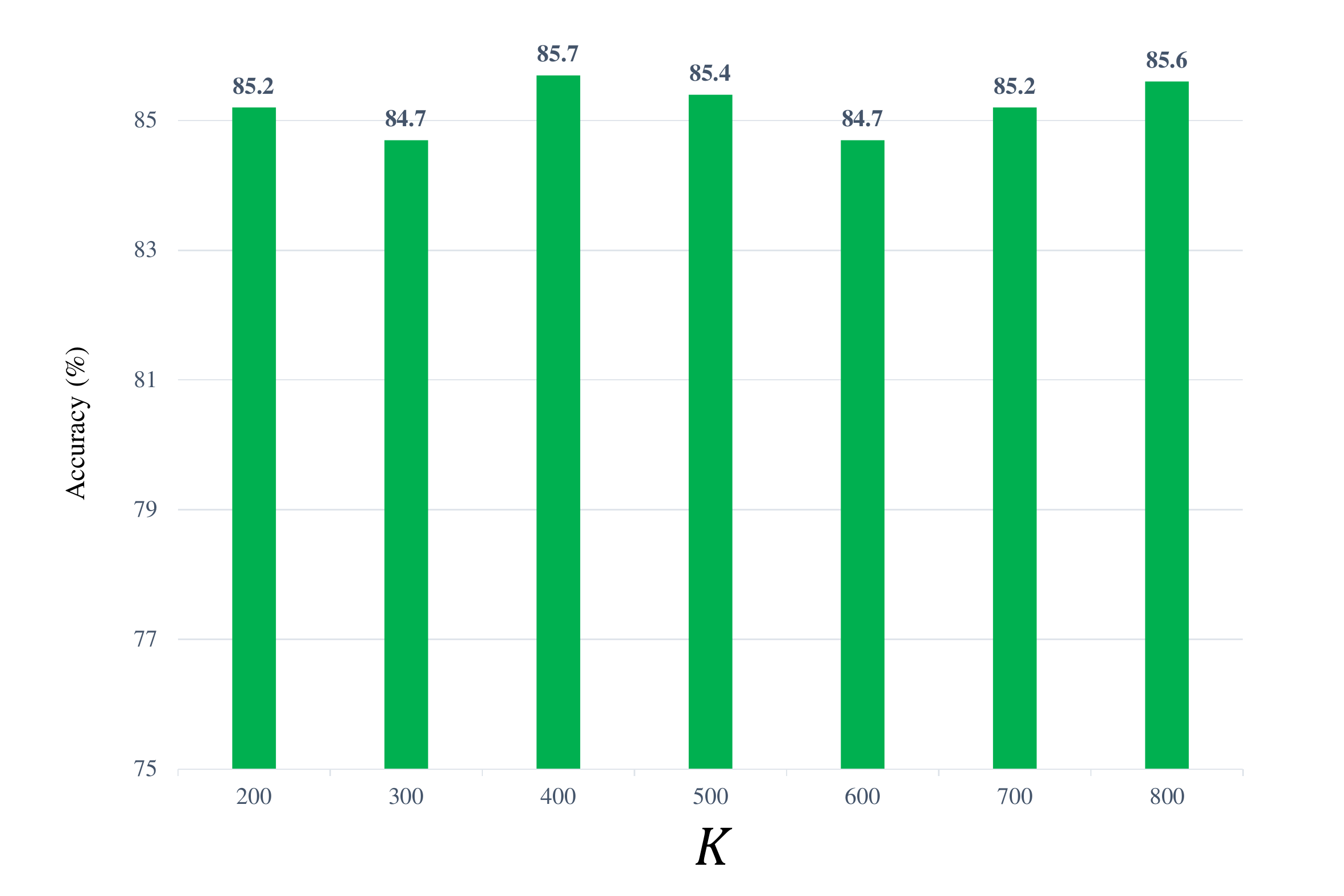}
    	\end{center}
    % 	\caption{}
    	\label{figure: K}
    \end{minipage}%
    }%
    \caption{The influence of parameters}
    \label{figure: alpha_beta_delta_K}
\end{figure}

\textbf{$i)$}
To demonstrate the efficiency of the dynamic label, we remove the process for obtaining dynamic labels and replace \begin{scriptsize}$\mathbf{F}$\end{scriptsize} with the fixed one-hot label matrix, where \begin{scriptsize}$\mathbf{E} \in \mathbb{R}^{C \times N_{l}}$\end{scriptsize} denotes the label matrix, $N_l$ represents the number of labeled training data. The objective function can be rewritten as:
\begin{equation}
\scriptsize
\begin{split}
        &\mathop {\arg \min}\limits_{\mathbf{D},\mathbf{S},\mathbf{B}}     
        \mathcal{F}(\mathbf{D},\mathbf{S},\mathbf{B})\\
		&=\left\| \mathbf{X} - \mathbf{D}\mathbf{S} \right\|_F^2 
		+ 2\alpha \left\| \mathbf{S} \right\|_{\ell_1} 
		+ \beta \left\| \mathbf{E} - \mathbf{B}\mathbf{S_{l}} \right\|_F^2
		+ \delta {\kern 2pt} \text{tr} \left( \mathbf{\Delta}_l \mathbf{S}^T \mathbf{S} \right)\\
		&{\rm{s}}.t.\left\| {{{\mathbf{d}}_{k}}} \right\|_2^2 \le 1, {\kern 4pt} \left\| {{{\mathbf{b}}_{k}}} \right\|_2^2 \le 1 {\kern 4pt} \left( {k = 1,2, \cdots K} \right)\\
\end{split}
\label{equation: Objective_function_fixed_label}
\end{equation}
where \begin{scriptsize}$\mathbf{S}_l \in \mathbb{R}^{K \times N_{l}}$\end{scriptsize} denotes the labeled parts of \begin{scriptsize}$\mathbf{S}$\end{scriptsize}. Figure~\ref{figure: dy-lb} shows the experimental results. From this figure, we find that the dynamic label term has a significant influence on the performance of the two datasets.

\textbf{$ii)$}
In the following experiments, we evaluate the influences of parameters on the UCM-LU dataset.  Figure~\ref{figure: alpha_beta_delta_K} shows the experimental results. In general experience, there are mainly four parameters (e.g. $\alpha$, $\beta$, $\delta$, $K$) affect the experimental results. More specifically, $\beta$ influences the performance independently. Thus we fix $\alpha$, $\delta$ and $K$ to observe the effect. From the figure, we can see that our method is sensitive to $\beta$. For $\alpha$ and $\delta$, they interact with each other. Thus we fix $\beta$ and $K$ to observe the impact of these two parameters simultaneously. We find that our method has strong adaptability to the two parameters, thus we can flexibly  choose the pairwise $\alpha$ and $\delta$ to obtain a good performance. Following, we evaluate the influence of dictionary size $K$. In this experiment, we tune the parameters $\alpha$, $\beta$, $\delta$ to obtain the optimal performance for each dictionary size. From the figure, we can conclude that the proposed DLDL approach is not sensitive to the dictionary size. Thus, we would choose a small size to reduce training time.

\section{Conclusion}
Previous label embedding dictionary learning based methods are not applicable in semi-supervised and unsupervised learning. Thus we propose a Dynamic Label Dictionary Learning (DLDL) algorithm to generate the soft label matrix for unlabeled data. The outstanding performance on two remote sensing datasets has demonstrated the efficiency of the proposed DLDL approach. It would be interesting for future work to expand the dynamic label to other subspace learning tasks.

\vfill\pagebreak

\bibliographystyle{IEEEbib.bst}
\bibliography{refs.bib}

\end{document}